\documentclass[sigconf]{acmart}

\usepackage{booktabs} 
\usepackage{mathtools}
\usepackage{algorithm}
\usepackage{multirow}
\usepackage{algpseudocode}
\usepackage{placeins}

\setcopyright{rightsretained}

\acmPrice{15.00}

\begin{document}
\title{Large-Scale Learnable Graph Convolutional Networks}

\author{Hongyang Gao}
\affiliation{%
  \institution{Washington State University}
  \city{Pullman}
  \state{WA}
  \postcode{99164}
}
\email{hongyang.gao@wsu.edu}

\author{Zhengyang Wang}
\affiliation{%
  \institution{Washington State University}
  \city{Pullman}
  \state{WA}
  \postcode{99164}
}
\email{zwang6@eecs.wsu.edu}

\author{Shuiwang Ji}
\affiliation{%
  \institution{Washington State University}
  \city{Pullman}
  \state{WA}
  \postcode{99164}
}
\email{sji@eecs.wsu.edu}

\renewcommand{\shortauthors}{H. Gao et al.}

\copyrightyear{2018} 
\acmYear{2018} 
\setcopyright{acmcopyright}
\acmConference[KDD '18]{The 24th ACM SIGKDD International Conference on Knowledge Discovery \& Data Mining}{August 19--23, 2018}{London, United Kingdom}
\acmBooktitle{KDD '18: The 24th ACM SIGKDD International Conference on Knowledge Discovery \& Data Mining, August 19--23, 2018, London, United Kingdom}
\acmPrice{15.00}
\acmDOI{10.1145/3219819.3219947}
\acmISBN{978-1-4503-5552-0/18/08}

\begin{abstract} Convolutional neural networks (CNNs) have achieved great
success on grid-like data such as images, but face tremendous challenges in
learning from more generic data such as graphs. In CNNs, the trainable local
filters enable the automatic extraction of high-level features. The
computation with filters requires a fixed number of ordered units in the
receptive fields. However, the number of neighboring units is neither fixed
nor are they ordered in generic graphs, thereby hindering the applications of
convolutional operations. Here, we address these challenges by proposing the
learnable graph convolutional layer (LGCL). LGCL automatically selects a fixed
number of neighboring nodes for each feature based on value ranking in order
to transform graph data into grid-like structures in 1-D format, thereby
enabling the use of regular convolutional operations on generic graphs. To
enable model training on large-scale graphs, we propose a sub-graph training
method to reduce the excessive memory and computational resource requirements
suffered by prior methods on graph convolutions. Our experimental results on
node classification tasks in both transductive and inductive learning settings
demonstrate that our methods can achieve consistently better performance on
the Cora, Citeseer, Pubmed citation network, and protein-protein interaction
network datasets. Our results also indicate that the proposed methods using
sub-graph training strategy are more efficient as compared to prior
approaches. \end{abstract}

%
%



\maketitle

\section{Introduction}

Deep learning methods are becoming increasingly powerful in solving
various challenging artificial intelligence tasks. Among these deep
learning methods, convolutional neural
networks~(CNNs)~\cite{lecun1998gradient} have demonstrated promising
performance in many image-related applications, such as image
classification~\cite{imagenet_cvpr09}, semantic
segmentation~\cite{chen2016deeplab}, and object
detection~\cite{ren2015faster,he2017mask}. A variety of CNN models
have been proposed to continuously set the performance
records~\cite{krizhevsky2012imagenet,simonyan2014very,szegedy2015going,he2016deep}.
In addition to images, CNNs have also been successfully applied to
natural language processing tasks such as neural machine
translation~\cite{cho2014properties,luong2015effective,gehring2016convolutional}.
One common characteristic behind these tasks is that the data can be
represented by grid-like structures. This enables the use of
convolutional operations in the form of the same local filters
scanning every position on the input. Unlike traditional
hand-crafted filters, the local filters used in convolutional layers
are trainable. The networks can automatically decide what kind of
features to extract by learning the weights in these trainable
filters, thereby avoiding hand-crafted feature
extraction~\cite{wang2012end}.

\begin{figure}[t] \includegraphics[width=0.8\columnwidth]{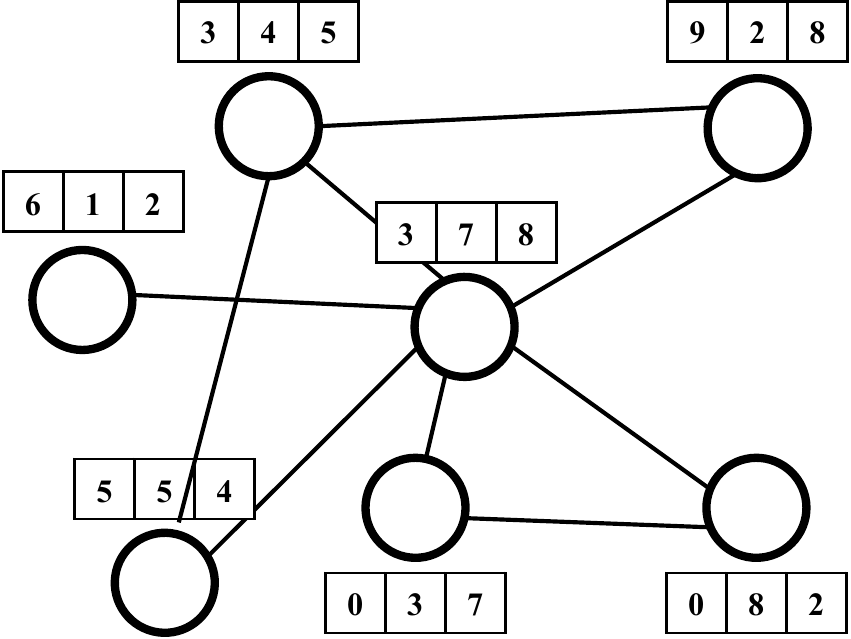}
\caption{An illustration of graph data. There are 7 nodes in this
graph and each node has 3 features. Each node in this graph may have
a different number of neighboring nodes, and there is no relative
order among them.} \label{fig:graph}
\end{figure}

In many real-world applications, the data can be naturally
represented as graphs, such as social, citation, and biological
networks. Figure~\ref{fig:graph} provides an illustration of graph
data. Many interesting discoveries can be made by analyzing these
graph data, such as social network
analysis~\cite{grover2016node2vec}. An important task on graph data
is node classification~\cite{kipf2016semi,velivckovic2017graph}, in
which models make predictions for every node in a graph based on
node features and graph topology. As mentioned above, CNNs, with the
power of automatic feature extraction, have achieved great success
on tasks with grid-like data, which can be considered as special
cases of graph data. Therefore, applying deep learning models,
especially CNNs, on graph tasks is appealing. However, using regular
convolutional operations on generic graphs faces two main
challenges. These challenges are resulted from the fact that regular
convolutions require the number of neighboring nodes for each node
remains the same, and these neighboring nodes are ordered. In
generic graphs, the numbers of neighboring nodes usually differ for
different nodes in a graph. In addition, among the neighboring nodes
of a node, there is no ranking information based on which we can
order them to ensure the output is deterministic. In this work, we
analyze the necessity of having a fixed number of ordered
neighboring nodes in regular convolutional operations and propose
elegant solutions to address these challenges.

Several recent studies tried to apply convolutional operations on
generic graphs. Graph convolutional
networks~(GCNs)~\cite{kipf2016semi} proposed to use a
convolution-like operation to aggregate features of all adjacent
nodes for each node, followed by a linear transformation to generate
new a feature representation for a given node. Specifically, all
feature vectors in the neighborhood, including the feature vector of
the central node itself, are summed up, weighted by non-trainable
weights depending on the number of neighbors. This can be thought of
as a convolution-like operation which, however, is intrinsically
different from the regular convolutional operation in two aspects.
First, it does not use the same local filter to scan every node;
that is, nodes that have different numbers of adjacent nodes have
filters of different sizes and weights. Second, the weights in the
filters are the same for all neighboring nodes in the receptive
field as they are determined by the number of neighbors.
Consequently, the weights are not learned. Graph attention
networks~(GATs)~\cite{velivckovic2017graph} employed the attention
mechanism~\cite{bahdanau2014neural} to obtain different and
trainable weights for adjacent nodes by measuring the correlation
between their feature vectors and that of the central node. Yet
graph attention operation still differs from the regular convolution
which learns weights in local filters directly. Moreover, the
attention mechanism requires extra computation in terms of pairs of
feature vectors, resulting in excessive memory and computational
resource requirements in practice.

In this work, we make two major contributions to applying CNNs on
generic graph data. First, we propose the learnable graph
convolutional layer (LGCL) to enable the use of regular
convolutional operations on graphs. Note that prior studies modified
the original convolutional operations to fit them for graph data. In
contrast, our LGCL transforms the graphs to enable the use of
regular convolutions. Our models based on LGCL achieve better
performance on both transductive learning and inductive node
classification tasks, as demonstrated by our experimental results.
Second, we observe another limitation of prior methods; that is,
their training process takes the adjacency matrix of the whole graph
as an input. This requires excessive memory and computational
resources when the graph has a large amount of nodes, which is
usually the case in real-world tasks. In order to overcome this
limitation, we develop a sub-graph training method, which is a
simple yet effective approach to allow the training of deep learning
methods on large-scale graph data. The sub-graph training method can
significantly reduce the amount of required memory and computational
resources, with negligible loss in terms of model performance.

\section{Related Work}\label{sec:related}

\begin{figure*}[pt!] \includegraphics[width=\textwidth]{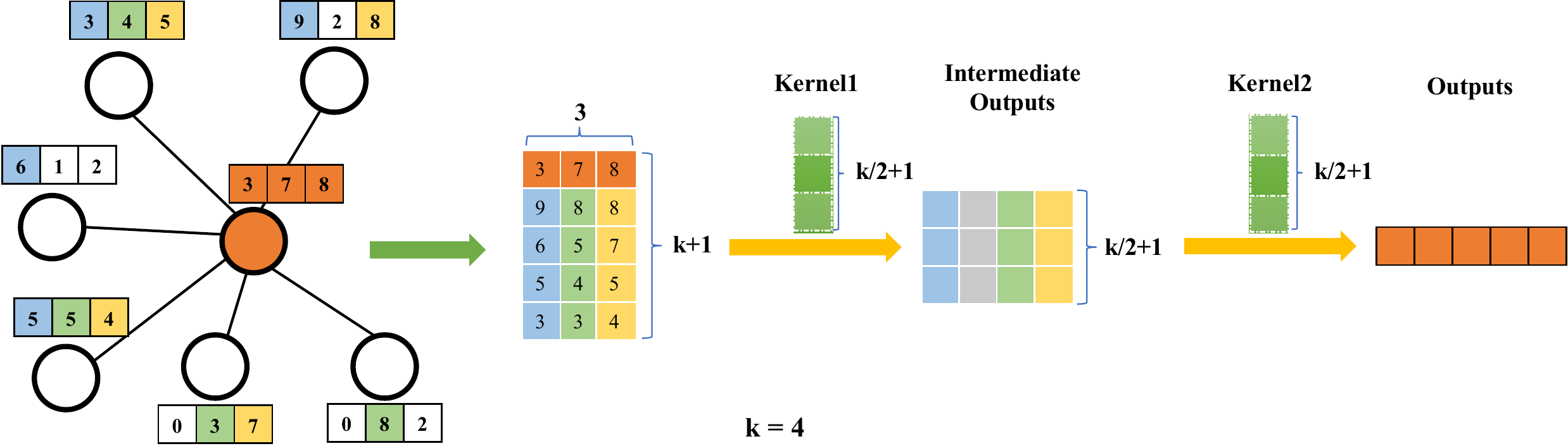}
\caption{An illustration of a learnable graph convolutional
layer~(LGCL). We consider a node with 6 adjacent nodes. Each node
has three features, represented by a 3-component feature vector.
This layer selects $k=4$ nodes in the neighborhood and employs a 1-D
CNN to produce a new vector representation of five features for the
central node, color-coded in orange. The left part describes the
process of selecting the $k$-largest values for each feature from
neighboring nodes. It can be seen from the graph that there are 6
neighbors. Since $k=4$, for each feature, four largest values are
selected from the neighborhood based on the ranking. For example,
the results of this selection process for the first feature is \{9,
6, 5, 3\} out of \{9, 6, 5, 3, 0, 0\}. By repeating the same process
for the other two features, we obtain $(k+1)$ 3-component feature
vectors, including that of the orange node itself. Concatenating
them gives a 1-D data of grid-like structure, which has $(k+1)$
positions and 3 channels. Afterwards, a 1-D CNN is applied to
generate the final feature vector. Specifically, we use two
convolutional layers with a kernel size of $(k/2+1)$ and without
padding. The numbers of output channels are 4 and 5, respectively.
In practice, the 1-D CNN can be any CNN model, as long as the final
output is a vector, serving as the new feature representation of the
central node.} \label{fig:layer}
\end{figure*}

A few recent studies have tried to apply convolutional operations on
graph data. Graph convolutional networks~(GCNs) were introduced
in~\cite{kipf2016semi} and achieved the state-of-art performance on
several node classification tasks. The authors defined and used a
convolution-like operation termed the spectral graph convolution.
This enables CNNs to directly operate on graphs. Basically, each
layer in GCNs updates the feature vector representation of each node
in the graph by considering the features of neighboring nodes. To be
specific, the layer-wise forward-propagation operation of GCNs can
be expressed as
\begin{equation}
\begin{aligned}
  X_{l+1} =
  \sigma(\hat{D}^{-\frac{1}{2}}\hat{A}\hat{D}^{-\frac{1}{2}}X_{l}W_{l}),
\end{aligned}\label{eq:gcn}
\end{equation}
where $X_{l}$ and $X_{l+1}$ are the input and output matrices of
layer $l$, respectively. For both matrices, the numbers of rows are
the same, corresponding to the number of nodes in the graph, while
the numbers of columns can be different, depending on the dimensions
of the input and output feature space. In Eq~(\ref{eq:gcn}),
$\hat{A} = A + I$ is used to aggregate feature vectors of adjacent
nodes, where $A$ is the adjacency matrix of the graph, and $I$ is
the identity matrix. Also, $\hat{A}$ is used, instead of $A$,
because the layers need to add self-loop connections to make sure
that the old feature vector of the node itself is taken into
consideration when updating the representation of a node. $\hat{D}$
is the diagonal node degree matrix, which is used to normalize
$\hat{A}$ so that the scale of feature vectors after aggregation
remains the same. $W_{l}$ is a trainable weight matrix and
represents a linear transformation that changes the dimension of
feature space. Therefore, the dimension of $W^{l}$ depends on how
many features that each node in the input and output have,
\emph{i.e.,} the number of columns in $X_{l}$ and $X_{l+1}$,
respectively. $\sigma(\cdot)$ denotes an activation function like
ReLU.

We analyze the convolution-like operation, which is the feature
aggregation step through pre-multiplying $X_{l}$ by
$\hat{D}^{-\frac{1}{2}}\hat{A}\hat{D}^{-\frac{1}{2}}$. Consider a
node with a feature vector corresponding to the $i$-th row in
$X_{l}$. The aggregation output, controlled by the $i$-th row in
$\hat{D}^{-\frac{1}{2}}\hat{A}\hat{D}^{-\frac{1}{2}}$, is a weighted
sum of the feature vectors of all of its adjacent nodes, including
the node itself. We can see that the operation is equivalent to
having a local filter for each node, whose receptive field consists
of the node itself and all its neighboring nodes. As is common that
nodes in a generic graph have different numbers of adjacent nodes,
the receptive field size varies, resulting in different local
filters. This is a key difference from the regular convolutional
operation, where the same local filter is applied to scan each
position in grid-like data. Moreover, while using local filters of
different sizes for graph data seems reasonable, it is worth noting
that there is no trainable parameter in
$\hat{D}^{-\frac{1}{2}}\hat{A}\hat{D}^{-\frac{1}{2}}$. In addition,
each adjacent node receives the same weight in the weighted sum,
which makes it a simple average. While CNNs achieve the power of
automatic feature extraction by learning the weights in local
filters, this non-trainable aggregation operation in GCNs limits the
capability of CNNs on generic graph data.

From this perspective, graph attention
networks~(GATs)~\cite{velivckovic2017graph} tried to enable
learnable weights when aggregating neighboring feature vectors by
employing the attention
mechanism~\cite{bahdanau2014neural,vaswani2017attention}. Like GCNs,
each node still has a local filter with a receptive field covering
the node itself and all of its adjacent nodes. When performing the
weighted sum of feature vectors, each neighbor receives a different
weight by measuring the correlation between its feature vector and
that of the central node. Mathematically, for a node $i$ and one of
its adjacent nodes $j$, the correlation measurement process between
layer $l$ and $l+1$ is given by
\begin{equation}
\begin{aligned}
  e_{l}^{i,j} &= a_{l}(W_{l} x_{l}^{i}, W_{l} x_{l}^{j}) \\
  \alpha_{l}^{i,j} &= \mbox{softmax}(e_{l}^{i,j}),
\end{aligned}\label{eq:gat}
\end{equation}
where $x_{l}^{i}$ and $x_{l}^{j}$ represent the corresponding
feature vectors, \emph{i.e.,} the $i$-th and $j$-th row in $X_{l}$,
respectively, $W_{l}$ is a shared linear transformation and $a_{l}$
represents a single-layer feed-forward neural network,
$\alpha_{l}^{i,j}$ is the weight for node $j$ in the feature
aggregation operation of node $i$. Although in this way, GATs
provide different and trainable weights to different adjacent nodes,
the learning process differs from that of regular CNNs where weights
in local filters are learned directly. Also, the attention mechanism
requires extra computation between a node and all of its adjacent
nodes, which will cause memory and computational resource problems
in practice.

Unlike these prior models, which modified the regular convolutional
operations to fit them for generic graph data, we instead propose to
transform graphs into grid-like data to enable the use of CNNs
directly. This idea was previously explored
in~\cite{niepert2016learning}. However, the transformation
in~\cite{niepert2016learning} is implemented in the preprocessing
process while our method includes the transformation in the
networks. Additionally, we introduce a sub-graph training method in
this work, which is a simple yet effective approach to allow
large-scale training.

\begin{figure*}[t] \includegraphics[width=\textwidth]{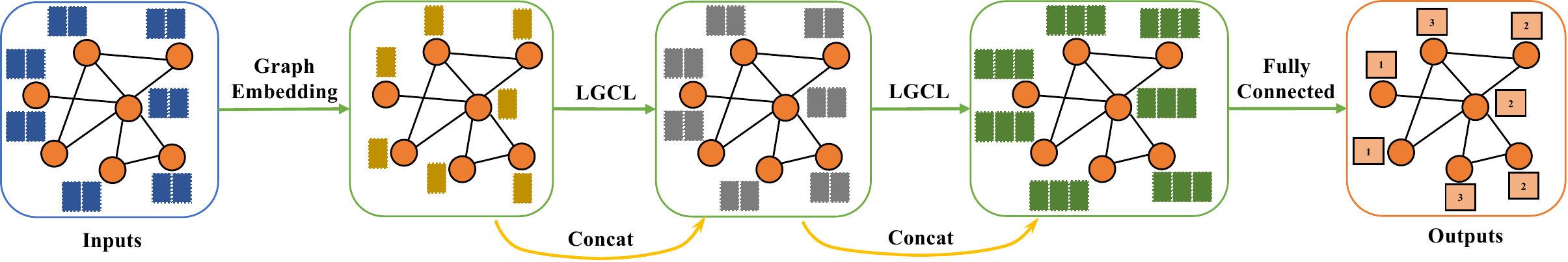}
\caption{An illustration of the proposed learnable graph
convolutional network~(LGCN). In this example, the nodes in the
input have two features. The input feature vectors are transformed
into low-dimensional representations using a graph embedding layer.
After that, we stack two LGCL layers with skip concatenation
connections to refine the feature vectors of each node. Finally, a
fully-connected layer is used for node classification. There are
three different classes in this example.} \label{fig:model}
\end{figure*}

\section{Methods}

In this section, we introduce the learnable graph convolutional
layer~(LGCL) and the sub-graph training strategy on generic graph
data. Based on these developments, we propose the large-scale
learnable graph convolutional networks~(LGCNs).

\subsection{Challenges of Applying Convolutional Operations on Graph Data}\label{sec:challenge}

In order to apply regular convolutional operations on graphs, we
need to overcome two main challenges that are caused by two major
differences between generic graphs and grid-like data. First, the
number of adjacent nodes usually varies for different nodes in a
generic graph. Second, we cannot order the neighboring nodes in
generic graphs, since there is no ranking information among them.
For example, in a social network, each person in the network can be
seen as a node and the edges represent friendships between people.
Obviously, the number of adjacent nodes differs for each node since
people can have different numbers of friends. Meanwhile, it is hard
to order these friends without additional information for ranking.

Note that grid-like data can be viewed as a special type of graph
data, where each node has a fixed number of ordered neighbors. As
convolutional operations apply directly on grid-like data such as
images, we analyze why the two characteristics mentioned above are
necessary to performing regular convolutions. To see the need of
having a fixed number of adjacent nodes with ranking information,
consider a convolutional filter with a size of $3 \times 3$ scanning
an image. We think of the image as a special graph by thinking of
each pixel as a node. During the scan, the computation involves a
central node with $3 \times 3-1=8$ adjacent nodes each time. These
$8$ nodes become neighbors of the central node by having edges
connecting them in the special graph. Meanwhile, we can order these
neighboring nodes by their relative positions with respect to the
central node. This is crucial to convolutional operations since the
correspondence between weights in the filter and nodes in the graph
must be maintained during the scan. For instance, in the example
above, the upper left weight in the $3 \times 3$ filter should
always be multiplied with the neighboring node at the top left of
the central node. Without such ranking information, the outputs of
convolution operations are no longer deterministic. We can see from
the above discussions that it is challenging to directly apply
regular convolutional operations on generic graph data. To address
these two challenges, we propose an approach to transform generic
graphs into grid-like data.

\subsection{Learnable Graph Convolutional Layers}

To enable the use of regular convolutional operations on generic
graphs, we propose the learnable graph convolutional layer~(LGCL).
Following the notations defined in Section~\ref{sec:related}, the
layer-wise propagation rule of LGCL is formulated as
\begin{equation}
\begin{aligned}
  &\tilde X_l = g(X_{l}, A, k),\\
  &X_{l+1} = c(\tilde X_{l}),
\end{aligned}\label{eq:lgcl}
\end{equation}
where $A$ is the adjacency matrix, $g(\cdot)$ is an operation that
performs the $k$-largest node selection to transform generic graphs
to data of grid-like structures, and $c(\cdot)$ denotes a regular
1-D CNN that aggregates neighboring information and outputs a new
feature vector for each node. We discuss $g(\cdot)$ and $c(\cdot)$
separately below.

\textbf{$k$-largest Node Selection.} We propose a novel method known
as the $k$-largest node selection to achieve the transformation from
graphs to grid-like data, where $k$ is a hyper-parameter of LGCL.
After this operation, each node aggregates neighboring information
and is represented in a 1-D grid-like format with $(k+1)$ positions.
The transformed data is then fed into a 1-D CNN to generate the
updated feature vector.

Suppose $X_{l} \in \mathbb{R}^{N \times C}$ with row vectors $x_l^1,
x_l^2, \cdots, x_l^N$, representing a graph of $N$ nodes where each
node has $C$ features. We are given the adjacency matrix $A \in
\mathbb{N}^{N \times N}$ and a fixed $k$. Now consider a specific
node $i$ whose feature vector is $x_l^i$ and it has $n$ neighboring
nodes. Through a simple look-up operation in $A$, we can obtain the
indices of these adjacent nodes, say $i_1, i_2, \cdots, i_n$.
Concatenating the corresponding feature vectors $x_l^{i_1},
x_l^{i_2}, \cdots, x_l^{i_n}$ outputs a matrix $M_l^i \in
\mathbb{R}^{n \times C}$. Without the loss of generalization, assume
that $n \geq k$. If $n < k$ in practice, we can pad $M_l^i$ using
columns of zeros. The $k$-largest node selection is conducted on
$M_l^i$; that is, for each column, we rank the $n$ values and select
$k$-largest values. This gives us a $k \times C$ output matrix. As
the columns in $M_l^i$ represent features, the operation is
equivalent to selecting $k$-largest values for each feature. By
inserting $x_l^i$ in the first row, the output becomes $\tilde M_l^i
\in \mathbb{R}^{(k+1) \times C}$. This is illustrated in the left
part of Figure~\ref{fig:layer}. By repeating this process for each
node, $g(\cdot)$ transforms $X_{l}$ to $\tilde X_l \in \mathbb{R}^{N
\times (k+1) \times C}$.

Note that $\tilde X_l$ can be viewed as a 1-D grid-like structure by
considering $N$, $(k+1)$, and $C$ as the batch size, the spatial
size, and the number of channels, respectively. Therefore, the
$k$-largest node selection function $g(\cdot)$ successfully achieves
the transformation from generic graphs to grid-like data. The
operation makes use of the natural ranking information among real
numbers and forces each node to have a fixed number of ordered
neighbors.

\textbf{1-D Convolutional Neural Networks.} As discussed in
Section~\ref{sec:challenge}, regular convolutional operations can be
directly applied on grid-like data. As $\tilde X_l \in \mathbb{R}^{N
\times (k+1) \times C}$ is 1-D, we employ a 1-D CNN model
$c(\cdot)$. The basic functionality of LGCL is to aggregate adjacent
information and update the feature vector for each node.
Consequently, it requires $X_{l+1} \in \mathbb{R}^{N \times D}$,
where $D$ is the dimension of the updated feature space. The 1-D CNN
$c(\cdot)$ should take $\tilde X_l \in \mathbb{R}^{N \times (k+1)
\times C}$ as input and output a matrix of dimension $N \times D$,
or equivalently, $N \times 1 \times D$. Basically, $c(\cdot)$
reduces the spatial size from $(k+1)$ to 1.

Note that $N$ is considered as the batch size, which is not related
to the design of $c(\cdot)$. As a result, we focus on only one data
sample, \emph{i.e.,} one node in the graph. Taking the example
above, for node $i$, the transformed output is $\tilde M_l^i \in
\mathbb{R}^{(k+1) \times C}$, which serves as the input to
$c(\cdot)$. Due to the fact that any regular convolutional operation
with a filter size larger than one and no padding reduces the
spatial size, the simplest $c(\cdot)$ has only one convolutional
layer with a filter size of $(k+1)$ and no padding. The numbers of
input and output channels are $C$ and $D$, respectively. Meanwhile,
any multi-layer CNN can be employed, provided its final output has
the dimension of $1\times D$. The right part of
Figure~\ref{fig:layer} illustrates an example of a two-layer CNN.
Again, applying $c(\cdot)$ for all the $N$ nodes outputs $X_{l+1}
\in \mathbb{R}^{N \times D}$. In summary, our LGCL transforms
generic graphs to grid-like data using the proposed $k$-largest node
selection and applies a regular 1-D CNN to perform feature
aggregation and refine the feature vector for each node.

\subsection{Learnable Graph Convolutional Networks}\label{sec:LGCN}

It is known that deeper networks usually yield better performance.
However, prior deep models on graphs like GCNs only have two layers.
While they suffer from performance loss when going
deeper~\cite{kipf2016semi}, our LGCL enables a deeper design,
resulting in the learnable graph convolutional networks~(LGCNs) for
graph node classification. We build LGCNs based on the architecture
of densely connected convolutional
networks~(DCNNs)~\cite{huang2016densely,he2015deep}, which achieved
state-of-the-art performance in the ImageNet classification
challenge~\cite{ImageNetCNN}.

\begin{algorithm}[t]
\caption{Sub-Graph Selection Algorithm}\label{algo:batch_select}
\begin{algorithmic}[1]
\floatname{algorithm}{Algorithm}
\renewcommand{\algorithmicrequire}{\textbf{Input:}}
\renewcommand{\algorithmicensure}{\textbf{Output:}}
\algnewcommand\AND{\textbf{and }}

\Require Adjacency matrix $A$, Number of nodes $N$, Sub-graph size
$N_s$,
         Initial number of nodes $N_{init}$,
         Maximum number of nodes expanded per iteration $N_{m}$
\Ensure A set of nodes $S$ as a sub-graph

\State S = $\phi$
\State initNodes = sample $N_{init}$ nodes from $N$ nodes.
\State S = S $\cup$ initNodes
\State newAddNodes = initNodes
\While{size(S) < $N_{s}$ \AND size(newAddNodes) $\ne$ 0}
\State candidateNodes = BFS(newAddNodes, A)

\Comment{Obtain first-order neighboring nodes of newAddNodes}
\State newAddNodes = candidateNodes $\setminus$  S

\If{size(newAddNodes) > $N_{m}$}
\State newAddNodes = sample $N_{m}$ nodes from newAddNodes
\EndIf

\If{size(newAddNodes) + size(S) > $N_s$}
\State $N_r$ = $N_s$ - size(S)
\State newAddNodes = sample $N_r$ nodes from newAddNodes
\EndIf

\State S = S $\cup$ newAddNodes

\EndWhile

\State \Return S

\end{algorithmic}
\end{algorithm}

\begin{figure}[t] \includegraphics[width=\columnwidth]{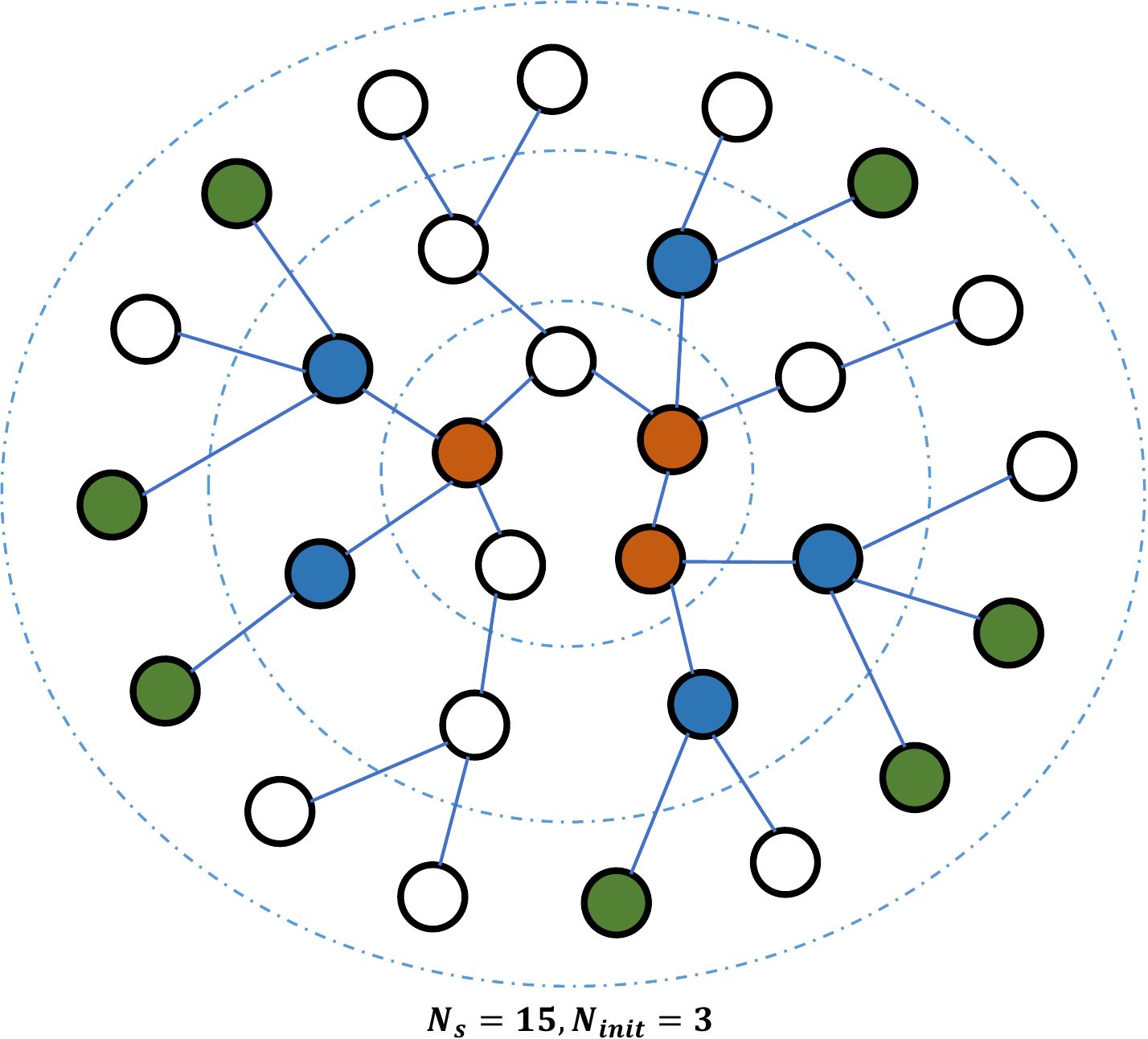}
\caption{An example of the sub-graph selection process. We start
with $N_{init}=3$ randomly sampled nodes and obtain a sub-graph of
$N_s = 15$ nodes. In the first iteration, we use BFS to find all the
first-order neighboring nodes of the 3 initial nodes (orange),
excluding themselves. Among these nodes, we randomly select $N_m=5$
nodes (blue). In the next iteration, we select $N_m=7$ nodes from
neighbors of the blue nodes, excluding previously selected nodes.
Note that $N_m$ changes for the two iterations, which is a flexible
choice in practice. After two iterations, we have selected
$3+5+7=15$ nodes and obtained a required sub-graph. These nodes,
along with the corresponding adjacency matrix, will form the input
to the LGCN in a training iteration.} \label{fig:batch}
\end{figure}

In LGCNs, we first apply a graph embedding layer to produce
low-dimensional representations of nodes, since the original inputs
are usually very high-dimensional feature vectors in some graph
dataset, such as the Cora~\cite{sen2008collective}. The graph
embedding layer is essentially a linear transformation in the first
layer expressed as
\begin{equation}
\begin{aligned}
  X_1 = X_0 W_0,
\end{aligned}\label{eq:embedding}
\end{equation}
where $X_0 \in \mathbb{R}^{N \times C_0}$ represents the
high-dimensional input and $W_0 \in \mathbb{R}^{C_0 \times C_1}$
changes the dimension of feature space from $C_0$ to $C_1$. As a
result, $X_1 \in \mathbb{R}^{N \times C_1}$ and $C_1 < C_0$.
Alternatively, a GCN layer can be used for graph embedding. As
illustrated in Section~\ref{sec:related}, the number of training
parameters in a GCN layer is equal to that of a regular graph
embedding layer.

After the graph embedding layer, we stack multiple LGCLs, according
to the complexity of the graph data. As each LGCL only aggregates
information from first-order neighboring nodes, \emph{i.e.,} direct
neighboring nodes, stacked LGCLs can collect information from a
larger set of nodes, which is commonly done in regular CNNs. In
order to promote the model performance and facilitate the training
process, we apply skip connections to concatenate the inputs and
outputs of LGCLs. Finally, a fully-connected layer is used before
the $\mbox{softmax}$ function for final predictions.

Following the design principle of LGCNs, $k$ and the number of
stacked LGCLs are the most important hyper-parameters. The average
degree of nodes in the graph can be a good reference for selecting
$k$. Meanwhile, the number of LGCLs should depend on the complexity
of tasks, such as the number of classes, the number of nodes in a
graph, etc. More complicated tasks require deeper models.

\begin{table*}[t]
\centering \caption{Summary of datasets used in our
experiments~\cite{yang2016revisiting,zitnik2017predicting}. The
Cora, Citeseer, and Pubmed datasets are used for transductive
learning experiments, while the PPI dataset is for inductive
learning experiments. The degree attribute listed is the average
node degree of each dataset, which helps the selection of the
hyper-parameter $k$ in LGCLs.} \label{table:datasets}
\begin{tabular}{  l   c  c  c  c  c  c  c  c }
    \hline
    \textbf{Dataset} & \textbf{\#Nodes} &
    \textbf{\#Features} & \textbf{\#Classes} & \textbf{\#Training Nodes} &
    \textbf{\#Validation Nodes} & \textbf{\#Test Nodes} & \textbf{Degree} & \textbf{Setting}  \\ \hline\hline
    Cora      & 2708   & 1433 & 7   & 140   & 500 & 1000 & 4  & Transductive \\ \hline
    Citeseer  & 3327   & 3703 & 6   & 120   & 500 & 1000 & 5  & Transductive \\ \hline
    Pubmed    & 19717  & 500  & 3   & 60    & 500 & 1000 & 6  & Transductive \\ \hline
    PPI       & 56944  & 50   & 121 & 44906 (20 graphs) & 6514 (2 graphs) & 5524 (2 graphs) & 31 & Inductive    \\
    \hline
\end{tabular}
\end{table*}

\subsection{Sub-Graph Training on Large-Scale Data}

Most prior deep models on graphs suffer from another limitation. In
particular, during training the inputs are the feature vectors of
all the nodes along with the adjacency matrix of the whole graph,
whose sizes become large for large graph data. These prior models
work properly on small-scale graphs. However, for large-scale
graphs, those methods usually result in excessive memory and
computational resource requirements, which limit the practical
applications of these models.

Similar problems also happen for deep neural networks on other types
of data, such as grid-like data. For example, deep models on image
segmentation usually use randomly cropped patches when dealing with
large images. Motivated by this strategy, we intend to randomly
``crop'' a graph to obtain smaller graphs for training. However,
while a rectangular patch of an image naturally maintains
neighboring information among pixels, how to handle irregular
connections between nodes in a graph remains challenging.

In this work, we propose a sub-graph selection algorithm to address
the memory and computational resource problems on large-scale graph
data, as shown in Algorithm~\ref{algo:batch_select}. Given a graph,
we first sample some initial nodes. Staring from them, we use the
Breadth-First-Search (BFS) algorithm to expand adjacent nodes into
the sub-graph iteratively. With multiple iterations, high-order
neighboring nodes of the initial nodes are included. Note that we
use a single parameter $N_m$ in Algorithm~\ref{algo:batch_select}
for simplicity. In practice, we can set $N_m$ to different values
for each iteration. Figure~\ref{fig:batch} provides an example of
the sub-graph selection process.

With such randomly ``cropped'' sub-graphs, we are able to train deep models on
large-scale graphs. In addition, we can take advantage of the mini-batch
training strategy to accelerate the learning process. In each training
iteration, we can use the proposed sub-graph selection algorithm to sample several
sub-graphs and put them in a mini-batch. The corresponding feature vectors and
adjacency matrices form the inputs to the networks.

\section{Experimental Studies}

In this section, we evaluate our proposed large-scale learnable
graph convolutional networks~(LGCNs) on node classification tasks
under both transductive and inductive learning settings. In addition
to comparisons with prior state-of-the-art models, some performance
studies are performed to investigate how to choose hyper-parameters.
Experiments are also conducted to analyze the training strategy
based on the proposed sub-graph selection algorithm. Experimental
results show that LGCNs yield improved performance, and the
sub-graph training is much more efficient than whole-graph training.
Our code is publicly
available\footnote{\url{https://github.com/divelab/lgcn/}}.

\subsection{Datasets}

In our experiments, we focus on node classification tasks under both
transductive and inductive learning settings.

\textbf{Transduction Learning.} Under the transductive setting, the
unlabeled testing data are accessible and available during training.
To be specific, for node classification, only a part of nodes in the
graph are labeled. The testing nodes, which are also in the same
graph, are accessible during training, including their features and
connections, except for the labels. This means the training process
knows about the graph structure that contains testing nodes. We use
three standard benchmark datasets for transductive learning
experiments; those are the Cora, Citeseer, and
Pubmed~\cite{sen2008collective}, as summarized in
Table~\ref{table:datasets}. These three datasets are citation
networks with nodes and edges representing documents and citations,
respectively. The feature vector of each node corresponds to a
bag-of-word representation for a document. For these three datasets,
we employ the same experimental settings as those in
GCN~\cite{kipf2016semi}. For each class, 20 nodes are used for
training, 500 nodes are used for validation and 1,000 nodes are used
for testing.

\textbf{Inductive Learning.} For inductive learning, the testing
data are not available during training, which means the training
process does not learn about the structure of test graphs. In
inductive learning tasks, we usually have different training,
validation, and testing graphs. During training, the model only use
the training graphs without access to validation and testing graphs.
We use the protein-protein interaction~(PPI)
dataset~\cite{zitnik2017predicting}, which contains 20 graphs for
training, 2 graphs for validation, and 2 graphs for testing. Since
the graphs for validation and testing are separate, the training
process does not use them. There are 2,372 nodes in each graph on
average. Each node has 50 features including positional, motif genes
and signatures. Each node has multiple labels from 121 classes.

\subsection{Experimental Setup}\label{sec:expsetup}

We describe the experimental setup under both transductive and
inductive learning settings.

\textbf{Transduction Learning.} In transductive learning tasks, we
employ the proposed LGCN models as illustrated in
Figure~\ref{fig:model}. Since transductive learning datasets employ
high-dimensional bag-of-word representations as feature vectors of
nodes, the inputs go through a graph embedding layer to reduce the
dimension. Here, we use a GCN layer as the graph embedding layer.
The dimension of the embedding output is 32. Then we apply LGCLs,
each of which uses $k=8$ and produces 8-component feature vectors.
For the Cora, Citeseer, and Pubmed, we stack 2, 1, and 1 LGCLs,
respectively. We use concatenation in skip connections. Finally, a
fully-connected layer is used as a classifier to make predictions.
Before the fully-connected layer, we perform a simple sum to
aggregate feature vectors of adjacent nodes.
Dropout~\cite{srivastava2014dropout} is applied on both input
feature vectors and adjacency matrices in each layer with rates of
0.16 and 0.999, respectively. All LGCN models in transductive
learning tasks use the sub-graph training strategy. The sub-graph
size is set to $2,000$.

\textbf{Inductive Learning.} For inductive learning, the same LGCN
model as above is used except for some hyper-parameters. For the
graph embedding layer, the dimension of output feature vectors is
128. We stack two LGCLs with $k=64$. We also employ the sub-graph
training strategy, with sub-graph initial node size equal to 500 and
200. Dropout with a rate of 0.9 is applied in each layer.

For both transductive and inductive learning LGCN models, the
following configurations are shared. For all layers, only the
identity activation function is used, which means no nonlinearity is
involved in the networks. In order to avoid over-fitting, the $L_2$
regularization with $\lambda=0.0005$ is applied. For training, the
Adam optimizer~\cite{kingma2014adam} with a learning rate of 0.1 is
used. Weights in LGCNs are initialized by the Glorot
initialization~\cite{glorot2010understanding}. We employ the early
stopping strategy based on the validation accuracy and train 1,000
epochs at most.

\begin{table}[t]
\centering \caption{Results of transductive learning experiments in
terms of node classification accuracies on the Cora, Citeseer, and
Pubmed datasets. LGCN$_{sub}$ denotes the LGCN model using the
sub-graph training strategy.} \label{table:trans}
\begin{tabular}{  l   c  c  c  }
    \hline
    \textbf{Models} & \textbf{Cora} & \textbf{Citeseer} & \textbf{Pubmed} \\ \hline\hline
    DeepWalk~\cite{perozzi2014deepwalk}            & 67.2\% & 43.2\%  & 65.3\%   \\ \hline
    Planetoid~\cite{yang2016revisiting}            & 75.7\% & 64.7\%  & 77.2\%   \\ \hline
    Chebyshev~\cite{defferrard2016convolutional}   & 81.2\% & 69.8\%  & 74.4\%   \\ \hline
    GCN~\cite{kipf2016semi}                        & 81.5\% & 70.3\%  & 79.0\%   \\ \hline
    \textbf{$\mathbf{LGCN}_{sub} \mathbf{(Ours)}$} & \textbf{83.3 $\pm$ 0.5\%}
                                                   & \textbf{73.0 $\pm$ 0.6\%}
                                                   & \textbf{79.5 $\pm$ 0.2\%} \\
    \hline
\end{tabular}
\end{table}

\begin{table}[t]
\centering \caption{Results of inductive learning experiments in
terms of micro-averaged F1 scores on the PPI dataset.}
\label{table:induc}
\begin{tabular}{  l   c }
    \hline
    \textbf{Models} & \textbf{PPI} \\ \hline\hline
    GraphSAGE-GCN~\cite{hamilton2017inductive}     & 0.500 \\ \hline
    GraphSAGE-mean~\cite{hamilton2017inductive}    & 0.598 \\ \hline
    GraphSAGE-pool~\cite{hamilton2017inductive}    & 0.600 \\ \hline
    GraphSAGE-LSTM~\cite{hamilton2017inductive}    & 0.612 \\ \hline
    \textbf{$\mathbf{LGCN}_{sub} \mathbf{(Ours)}$} & \textbf{0.772 $\pm$ 0.002} \\
    \hline
\end{tabular}
\end{table}

\subsection{Analysis of Results}

The experimental results are summarized in Tables~\ref{table:trans}
and~\ref{table:induc} for transductive and learning settings,
respectively.

\textbf{Transduction Learning.} For transductive learning
experiments, we report node classification accuracies as
in~\cite{kipf2016semi}. Table~\ref{table:trans} provides the
comparisons with other graph models. According to the results, our
LGCN models achieve better performance over the current
state-of-the-art GCNs by a margin of 1.8\%, 2.7\%, and 0.6\% on the
Cora, Citeseer, and Pubmed datasets, respectively.

\textbf{Inductive Learning.} For inductive learning experiments, we
report micro-averaged F1 scores like~\cite{hamilton2017inductive}.
From table~\ref{table:induc}, we can observe that our LGCN model
outperforms GraphSAGE-LSTM by a margin of 16\%. Without observing
the structure of test graphs in training, the LGCN model still
achieves good generalization.

The results above show that the proposed LGCN models on generic
graphs consistently yield new state-of-the-art performance in node
classification tasks on different datasets. These results
demonstrate the effectiveness of applying regular convolutional
operations on transformed graph data. In addition, the proposed
transformation approach through the $k$-largest node selection is
shown to be effective.

\subsection{LGCL versus GCN Layers}

It may be argued that our LGCN models employ a deeper network
architecture than GCNs, which could explain the improved
performance. However, the performance of GCNs is reported to
decrease when going deeper by stacking more layers. In addition, we
conduct another experiment by replacing all LGCLs in LGCN models by
GCN layers, denoted as LGCN$_{sub}$-GCN model. All the other
settings remain the same in order to ensure the fairness of the
comparisons. Table~\ref{table:lgcl_vs_gcn} provides the comparison
results between LGCN$_{sub}$ and LGCN$_{sub}$-GCN. The results show
that LGCN$_{sub}$ has better performance than LGCN$_{sub}$-GCN,
which indicates that the LGCL is more effective than the GCN layer.

\begin{table}[t]
\centering \caption{Results of transductive learning experiments for
comparing the LGCN$_{sub}$ and GCN layers on the Cora, Citeseer, and
Pubmed datasets. Using the network architecture of LGCN$_{sub}$, we
replace LGCLs by GCN layers, resulting in the LGCN$_{sub}$-GCN
model.} \label{table:lgcl_vs_gcn}
\begin{tabular}{  l   c  c  c}
    \hline
    \textbf{Models} & \textbf{Cora} & \textbf{Citeseer} & \textbf{Pubmed} \\ \hline\hline
    LGCN$_{sub}$-GCN   & 82.2 $\pm$ 0.5\%   & 71.1 $\pm$ 0.5\% & 79.0 $\pm$ 0.2\% \\ \hline
   \textbf{$\mathbf{LGCN}_{sub} \mathbf{(Ours)}$} & \textbf{83.3 $\pm$ 0.5\%}
                                                  & \textbf{73.0 $\pm$ 0.6\%}
                                                  & \textbf{79.5 $\pm$ 0.2\%} \\
    \hline
\end{tabular}
\end{table}

\subsection{Sub-Graph versus Whole-Graph Training}

For the experiments above, we use the sub-graph training strategy to
learn the LGCN models, which aims at saving memory and training
time. However, since the sub-graph selection algorithm samples some
nodes as a sub-graph from the whole graph, it means that the models
trained in this way do not learn about the structure of whole graph
during training. Meanwhile, in transductive learning tasks, the
information of testing nodes may be ignored, which raises the risk
of performance loss. To address this concern, we perform experiments
on transductive learning tasks to compare the sub-graph training
strategy with the previous whole-graph training strategy. Through
the experiments, we show the advantages of using the sub-graph
training strategy, with negligible loss in terms of model
performance.

For the sub-graph selection process described in
Algorithm~\ref{algo:batch_select}, the algorithm starts with some
initial nodes that are randomly selected. In transductive learning
tasks, we sample initial nodes only from the nodes with training
labels to make sure that training can be conducted. To be specific,
we sample 140, 120, and 60 initial nodes when selecting the
sub-graph for the Cora, Citeseer, and Pubmed datasets, respectively.
For each iteration in the sub-graph selection algorithm, we do not
set $N_m$ to limit the number of nodes expanded into the sub-graph.
The maximum number of nodes in the sub-graph is set to 2,000 for all
the three datasets, which is an feasible size for our GPUs in hand.

For comparison, we perform experiments using the same LGCN models,
but train them using the same whole-graph training strategy as GCNs,
which means the inputs are representations of the entire graph. We
denote such models as LGCN$_{whole}$, compared to LGCN$_{sub}$ with
the sub-graph training strategy. The comparing results of these two
models with GCNs are provided in Table~\ref{table:batch_vs_unbatch}.
The number of nodes reported represents how many nodes are used for
one iteration of training. The time reported here is the training
time for running 100 epochs using a single TITAN Xp GPU.

\begin{table}[t]
\centering \caption{Results of transductive learning experiments for
comparing the sub-graph training and whole-graph training strategies
on the Cora, Citeseer, and Pubmed datasets. For comparison, we
conduct experiments on LGCNs that employ the same whole-graph
training strategy as GCNs, denoted as LGCN$_{whole}$.}
\label{table:batch_vs_unbatch}
\tabcolsep=0.12cm
\begin{tabular}{  l  l  c  c  c }
    \hline
      &  & \textbf{Cora} & \textbf{Citeseer} & \textbf{Pubmed} \\ \hline\hline
    \multirow{3}{*}{\textbf{GCN}}
      & \# Nodes & 2708 & 3327 & 19717 \\ \cline{2-5}
      & Accuracy   & 81.5\% & 70.3\% & 79.0\% \\ \cline{2-5}
      & Time       & \textbf{7s} & 4s & 38s \\ \hline\hline
    \multirow{3}{*}{$\mathbf{LGCN}_{whole}$}
      & \# Nodes & 2708 & 3327 & 19717 \\ \cline{2-5}
      & Accuracy   & 83.8 $\pm$ 0.5\% & 73.0 $\pm$ 0.6\% & 79.5 $\pm$ 0.2\% \\ \cline{2-5}
      & Time       & 58s & 30s & 1080s \\ \hline\hline
    \multirow{3}{*}{$\mathbf{LGCN}_{sub}$}
      & \# Nodes & 644 & 442 & 354 \\ \cline{2-5}
      & Accuracy   & 83.3 $\pm$ 0.5\% & 73.0 $\pm$ 0.6\% & 79.5 $\pm$ 0.2\% \\ \cline{2-5}
      & Time       & 14s & \textbf{3.6s} & \textbf{2.6s} \\ \hline
\end{tabular}
\end{table}

It can be seen that the actual numbers of nodes in the training
sub-graph for the Cora, Citeseer, and Pubmed datasets are 644, 442,
and 354, respectively, which are far smaller than the maximum
sub-graph size of 2,000. This indicates that the nodes in the Cora,
Citeseer, and Pubmed datasets are sparsely connected. Specifically,
starting from several initial nodes with training labels, only a
small set of nodes will be selected by expanding neighboring nodes
to form connected sub-graphs. While these datasets are usually
considered as a single large graph, the whole graph is actually
composed of several separate sub-graphs that have no connection to
each other. The sub-graph training strategy takes advantage of this
fact and makes efficient use of the nodes with training labels.
Since only the initial nodes have training labels and all their
connectivity information is included in the selected sub-graphs, the
amount of information loss in the sub-graph training is minimized,
resulting in negligible performance loss. This is demonstrated by
comparing the node classification accuracies of LGCN$_{sub}$ and
LGCN$_{whole}$. According to the results, LGCN$_{sub}$ models only
have a subtle performance loss of 0.5\% on the Cora dataset, while
yielding the same performance on the Citeseer and Pubmed datasets,
as compared to the LGCN$_{whole}$ models.

\begin{figure}[t] \includegraphics[width=\columnwidth]{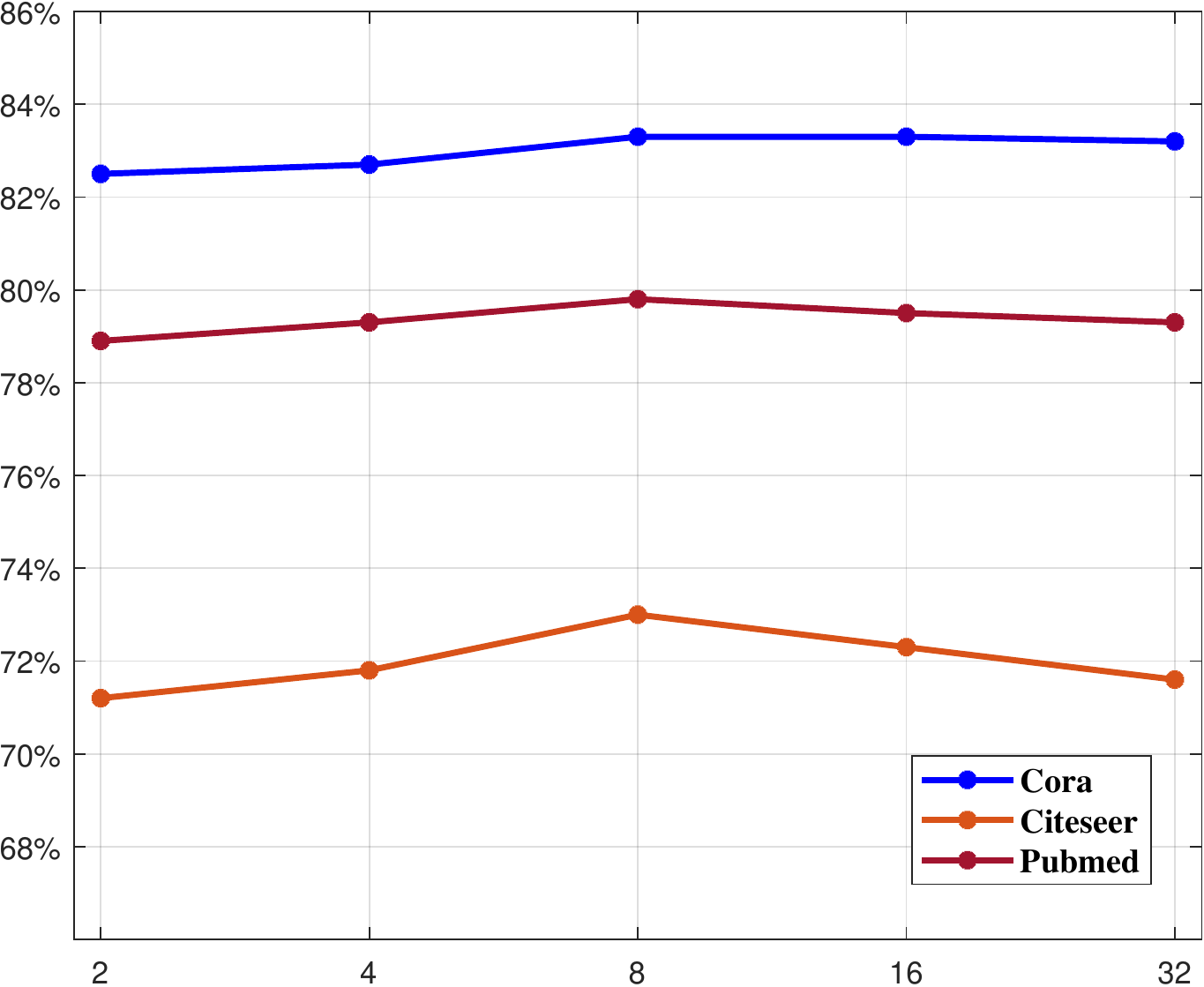}
\caption{Results of using different values of hyper-parameter $k$ in
LGCN models. On the Cora, Citeseer, and Pubmed datasets, we employ
the same experimental setups described in
Section~\ref{sec:expsetup}. We adjust the value of $k$ in
LGCN$_{sub}$ and report node classification accuracies in this
figure. It can be seen that $k=8$ achieves the best performance for
these datasets.} \label{fig:k_exp} \end{figure}

After investigating the risk of performance loss, we point out the
great advantages of the sub-graph training strategy in terms of
training speed. By using the sub-graph training, LGCN$_{sub}$ models
take a sub-graph of fewer nodes as inputs in contrast to the whole
graph, which is expected to greatly promote the training efficiency.
It can be seen from the results in
Table~\ref{table:batch_vs_unbatch} that the improvement is
outstanding. Although GCNs require simpler computation, its running
time is much longer than that of LGCN models on large-scale graph
datasets like the Pubmed. Powerful deep models are usually used on
large-scale data, which makes the sub-graph training strategy useful
in practice. The sub-graph training strategy enables using more
complex layers such as the proposed LGCLs without the concern of
long training time. As a result, our large-scale LGCNs with the
sub-graph training strategy are not only effective but also very
efficient.

\subsection{Performance Study of k}

As described in Section~\ref{sec:LGCN}, the average degree of nodes
in graph can be helpful when choosing the hyper-parameter $k$ in
LGCNs. In this part, we conduct experiments to show how different
values of $k$ affect the performance of LGCN models. We vary the
value of $k$ in LGCLs and observe the node classification accuracies
on the Cora, Citeseer, and Pubmed datasets. The values of $k$ are
selected from 2, 4, 8, 16, and 32, which cover a reasonable range of
integer values.

Figure~\ref{fig:k_exp} plots the performance change of LGCN models
under different values of $k$. As demonstrated in the figure, the
LGCN models achieve the best performance on all the three datasets
when choosing $k=8$. In the Cora, Citeseer, and Pubmed datasets, the
average node degrees are 4, 5, and 6, respectively. This indicates
that the best $k$ is usually a bit larger than the average node
degree in the dataset. When $k$ is too large, the performance of
LGCN models decreases. A possible explanation is that if $k$ is much
larger than the average node degree in the graph, too many zero
padding is used in the $k$-largest node selection process, which
compromises the performance of the following 1-D CNN models. For the
inductive learning task on the PPI dataset, we also explore
different values of $k$. The best performance is given by $k=64$
while the average node degree is 31. This is consistent with our
results above.

\section{Conclusions and Future Work}

In this work, we propose the learnable graph convolutional layer
(LGCL), which transforms generic graphs to data of grid-like
structures and enables the use of regular convolutional operations.
The transformation is conducted through a novel $k$-largest node
selection process, which uses the ranking between node feature
values. Based on our LGCL, we build deeper networks, known as
learnable graph convolutional networks~(LGCNs), for node
classification tasks on graphs. Experimental results show that the
proposed LGCN models yield consistently better performance than
prior methods under both transductive and inductive learning
settings. Our LGCN models achieve new state-of-the-art results on
four different datasets, demonstrating the effectiveness of LGCLs.

In addition, we propose a sub-graph selection algorithm, resulting
in the sub-graph training strategy, which can solve the problem of
excessive requirements for memory and computational resources on
large-scale graph data. With the sub-graph training, the proposed
LGCN models are both effective and efficient. Our experiments
indicate that the sub-graph training strategy brings a significant
advantage in terms of training speed, with a negligible amount of
performance loss. The new training strategy is very useful as it
enables the use of more complex models efficiently.

Based on this work, we discuss several possible directions for
future work. First, our methods mainly address the node
classification problems. In practice, many other interesting tasks
can be formulated as graph classification problems, where each graph
has a label. While they are similar to image classification tasks,
current graph convolutional methods, including ours, are not able to
perform down-sampling on graphs, like the pooling operations on
image data. We need a layer to reduce the number of nodes
effectively, which is necessary for graph classification. Second,
our methods are mainly applied to generic graph data like citation
networks. For other data like text, our methods may also be helpful,
since we can treat text data as graphs. We will explore these
directions in the future.

\begin{acks}
This work was supported in part by National Science Foundation
grants DBI-1641223, IIS-1633359 and Defense Advanced Research
Projects Agency grant N66001-17-2-4031.
\end{acks}

\bibliographystyle{ACM-Reference-Format}
\bibliography{deep}


\begin{thebibliography}{31}


\ifx \showCODEN    \undefined \def \showCODEN     #1{\unskip}     \fi
\ifx \showDOI      \undefined \def \showDOI       #1{#1}\fi
\ifx \showISBNx    \undefined \def \showISBNx     #1{\unskip}     \fi
\ifx \showISBNxiii \undefined \def \showISBNxiii  #1{\unskip}     \fi
\ifx \showISSN     \undefined \def \showISSN      #1{\unskip}     \fi
\ifx \showLCCN     \undefined \def \showLCCN      #1{\unskip}     \fi
\ifx \shownote     \undefined \def \shownote      #1{#1}          \fi
\ifx \showarticletitle \undefined \def \showarticletitle #1{#1}   \fi
\ifx \showURL      \undefined \def \showURL       {\relax}        \fi
\providecommand\bibfield[2]{#2}
\providecommand\bibinfo[2]{#2}
\providecommand\natexlab[1]{#1}
\providecommand\showeprint[2][]{arXiv:#2}

\bibitem[\protect\citeauthoryear{Bahdanau, Cho, and Bengio}{Bahdanau
  et~al\mbox{.}}{2015}]%
        {bahdanau2014neural}
\bibfield{author}{\bibinfo{person}{Dzmitry Bahdanau},
  \bibinfo{person}{Kyunghyun Cho}, {and} \bibinfo{person}{Yoshua Bengio}.}
  \bibinfo{year}{2015}\natexlab{}.
\newblock \showarticletitle{Neural machine translation by jointly learning to
  align and translate}.
\newblock \bibinfo{journal}{\emph{International Conference on Learning
  Representations}} (\bibinfo{year}{2015}).
\newblock


\bibitem[\protect\citeauthoryear{Chen, Papandreou, Kokkinos, Murphy, and
  Yuille}{Chen et~al\mbox{.}}{2016}]%
        {chen2016deeplab}
\bibfield{author}{\bibinfo{person}{Liang-Chieh Chen}, \bibinfo{person}{George
  Papandreou}, \bibinfo{person}{Iasonas Kokkinos}, \bibinfo{person}{Kevin
  Murphy}, {and} \bibinfo{person}{Alan~L Yuille}.}
  \bibinfo{year}{2016}\natexlab{}.
\newblock \showarticletitle{Deeplab: Semantic image segmentation with deep
  convolutional nets, atrous convolution, and fully connected crfs}.
\newblock \bibinfo{journal}{\emph{Transactions on Pattern Analysis and Machine
  Intelligence}} (\bibinfo{year}{2016}).
\newblock


\bibitem[\protect\citeauthoryear{Cho, van Merri{\"e}nboer, Bahdanau, and
  Bengio}{Cho et~al\mbox{.}}{2014}]%
        {cho2014properties}
\bibfield{author}{\bibinfo{person}{Kyunghyun Cho}, \bibinfo{person}{Bart van
  Merri{\"e}nboer}, \bibinfo{person}{Dzmitry Bahdanau}, {and}
  \bibinfo{person}{Yoshua Bengio}.} \bibinfo{year}{2014}\natexlab{}.
\newblock \showarticletitle{On the Properties of Neural Machine Translation:
  Encoder--Decoder Approaches}.
\newblock \bibinfo{journal}{\emph{Syntax, Semantics and Structure in
  Statistical Translation}} (\bibinfo{year}{2014}), \bibinfo{pages}{103}.
\newblock


\bibitem[\protect\citeauthoryear{Defferrard, Bresson, and
  Vandergheynst}{Defferrard et~al\mbox{.}}{2016}]%
        {defferrard2016convolutional}
\bibfield{author}{\bibinfo{person}{Micha{\"e}l Defferrard},
  \bibinfo{person}{Xavier Bresson}, {and} \bibinfo{person}{Pierre
  Vandergheynst}.} \bibinfo{year}{2016}\natexlab{}.
\newblock \showarticletitle{Convolutional neural networks on graphs with fast
  localized spectral filtering}. In \bibinfo{booktitle}{\emph{Advances in
  Neural Information Processing Systems}}. \bibinfo{pages}{3844--3852}.
\newblock


\bibitem[\protect\citeauthoryear{Deng, Dong, Socher, Li, Li, and Fei-Fei}{Deng
  et~al\mbox{.}}{2009}]%
        {imagenet_cvpr09}
\bibfield{author}{\bibinfo{person}{J. Deng}, \bibinfo{person}{W. Dong},
  \bibinfo{person}{R. Socher}, \bibinfo{person}{L.-J. Li}, \bibinfo{person}{K.
  Li}, {and} \bibinfo{person}{L. Fei-Fei}.} \bibinfo{year}{2009}\natexlab{}.
\newblock \showarticletitle{{ImageNet: A Large-Scale Hierarchical Image
  Database}}. In \bibinfo{booktitle}{\emph{Proceedings of the IEEE Conference
  on Computer Vision and Pattern Recognition}}.
\newblock


\bibitem[\protect\citeauthoryear{Gehring, Auli, Grangier, and Dauphin}{Gehring
  et~al\mbox{.}}{2017}]%
        {gehring2016convolutional}
\bibfield{author}{\bibinfo{person}{Jonas Gehring}, \bibinfo{person}{Michael
  Auli}, \bibinfo{person}{David Grangier}, {and} \bibinfo{person}{Yann~N
  Dauphin}.} \bibinfo{year}{2017}\natexlab{}.
\newblock \showarticletitle{A convolutional encoder model for neural machine
  translation}.
\newblock \bibinfo{journal}{\emph{Annual Meeting of the Association for
  Computational Linguistics}} (\bibinfo{year}{2017}).
\newblock


\bibitem[\protect\citeauthoryear{Glorot and Bengio}{Glorot and Bengio}{2010}]%
        {glorot2010understanding}
\bibfield{author}{\bibinfo{person}{Xavier Glorot} {and} \bibinfo{person}{Yoshua
  Bengio}.} \bibinfo{year}{2010}\natexlab{}.
\newblock \showarticletitle{Understanding the difficulty of training deep
  feedforward neural networks}. In \bibinfo{booktitle}{\emph{Proceedings of the
  Thirteenth International Conference on Artificial Intelligence and
  Statistics}}. \bibinfo{pages}{249--256}.
\newblock


\bibitem[\protect\citeauthoryear{Grover and Leskovec}{Grover and
  Leskovec}{2016}]%
        {grover2016node2vec}
\bibfield{author}{\bibinfo{person}{Aditya Grover} {and} \bibinfo{person}{Jure
  Leskovec}.} \bibinfo{year}{2016}\natexlab{}.
\newblock \showarticletitle{node2vec: Scalable feature learning for networks}.
  In \bibinfo{booktitle}{\emph{Proceedings of the 22nd ACM SIGKDD international
  conference on Knowledge discovery and data mining}}. ACM,
  \bibinfo{pages}{855--864}.
\newblock


\bibitem[\protect\citeauthoryear{Hamilton, Ying, and Leskovec}{Hamilton
  et~al\mbox{.}}{2017}]%
        {hamilton2017inductive}
\bibfield{author}{\bibinfo{person}{William~L. Hamilton}, \bibinfo{person}{Rex
  Ying}, {and} \bibinfo{person}{Jure Leskovec}.}
  \bibinfo{year}{2017}\natexlab{}.
\newblock \showarticletitle{Inductive Representation Learning on Large Graphs}.
  In \bibinfo{booktitle}{\emph{NIPS}}.
\newblock


\bibitem[\protect\citeauthoryear{He, Gkioxari, Doll{\'a}r, and Girshick}{He
  et~al\mbox{.}}{2017}]%
        {he2017mask}
\bibfield{author}{\bibinfo{person}{Kaiming He}, \bibinfo{person}{Georgia
  Gkioxari}, \bibinfo{person}{Piotr Doll{\'a}r}, {and} \bibinfo{person}{Ross
  Girshick}.} \bibinfo{year}{2017}\natexlab{}.
\newblock \showarticletitle{Mask r-cnn}.
\newblock \bibinfo{journal}{\emph{IEEE International Conference on Computer
  Vision}} (\bibinfo{year}{2017}).
\newblock


\bibitem[\protect\citeauthoryear{He, Zhang, Ren, and Sun}{He
  et~al\mbox{.}}{2016a}]%
        {he2016deep}
\bibfield{author}{\bibinfo{person}{Kaiming He}, \bibinfo{person}{Xiangyu
  Zhang}, \bibinfo{person}{Shaoqing Ren}, {and} \bibinfo{person}{Jian Sun}.}
  \bibinfo{year}{2016}\natexlab{a}.
\newblock \showarticletitle{Deep residual learning for image recognition}. In
  \bibinfo{booktitle}{\emph{Proceedings of the IEEE conference on computer
  vision and pattern recognition}}. \bibinfo{pages}{770--778}.
\newblock


\bibitem[\protect\citeauthoryear{He, Zhang, Ren, and Sun}{He
  et~al\mbox{.}}{2016b}]%
        {he2015deep}
\bibfield{author}{\bibinfo{person}{Kaiming He}, \bibinfo{person}{Xiangyu
  Zhang}, \bibinfo{person}{Shaoqing Ren}, {and} \bibinfo{person}{Jian Sun}.}
  \bibinfo{year}{2016}\natexlab{b}.
\newblock \showarticletitle{Deep Residual Learning for Image Recognition}. In
  \bibinfo{booktitle}{\emph{Proceedings of the IEEE Conference on Computer
  Vision and Pattern Recognition}}.
\newblock


\bibitem[\protect\citeauthoryear{Huang, Liu, Weinberger, and van~der
  Maaten}{Huang et~al\mbox{.}}{2017}]%
        {huang2016densely}
\bibfield{author}{\bibinfo{person}{Gao Huang}, \bibinfo{person}{Zhuang Liu},
  \bibinfo{person}{Kilian~Q Weinberger}, {and} \bibinfo{person}{Laurens van~der
  Maaten}.} \bibinfo{year}{2017}\natexlab{}.
\newblock \showarticletitle{Densely connected convolutional networks}.
\newblock \bibinfo{journal}{\emph{IEEE Conference on Computer Vision and
  Pattern Recognition}} (\bibinfo{year}{2017}).
\newblock


\bibitem[\protect\citeauthoryear{Kingma and Ba}{Kingma and Ba}{2015}]%
        {kingma2014adam}
\bibfield{author}{\bibinfo{person}{Diederik Kingma} {and}
  \bibinfo{person}{Jimmy Ba}.} \bibinfo{year}{2015}\natexlab{}.
\newblock \showarticletitle{Adam: A method for stochastic optimization}.
\newblock \bibinfo{journal}{\emph{The International Conference on Learning
  Representations}} (\bibinfo{year}{2015}).
\newblock


\bibitem[\protect\citeauthoryear{Kipf and Welling}{Kipf and Welling}{2017}]%
        {kipf2016semi}
\bibfield{author}{\bibinfo{person}{Thomas~N Kipf} {and} \bibinfo{person}{Max
  Welling}.} \bibinfo{year}{2017}\natexlab{}.
\newblock \showarticletitle{Semi-supervised classification with graph
  convolutional networks}.
\newblock \bibinfo{journal}{\emph{International Conference on Learning
  Representations}} (\bibinfo{year}{2017}).
\newblock


\bibitem[\protect\citeauthoryear{Krizhevsky, Sutskever, and Hinton}{Krizhevsky
  et~al\mbox{.}}{2012a}]%
        {ImageNetCNN}
\bibfield{author}{\bibinfo{person}{Alex Krizhevsky}, \bibinfo{person}{Ilya
  Sutskever}, {and} \bibinfo{person}{Geoff Hinton}.}
  \bibinfo{year}{2012}\natexlab{a}.
\newblock \showarticletitle{ImageNet Classification with Deep Convolutional
  Neural Networks}.
\newblock In \bibinfo{booktitle}{\emph{Advances in Neural Information
  Processing Systems 25}}, \bibfield{editor}{\bibinfo{person}{P.~Bartlett},
  \bibinfo{person}{F.C.N. Pereira}, \bibinfo{person}{C.J.C. Burges},
  \bibinfo{person}{L.~Bottou}, {and} \bibinfo{person}{K.Q. Weinberger}} (Eds.).
  \bibinfo{pages}{1106--1114}.
\newblock


\bibitem[\protect\citeauthoryear{Krizhevsky, Sutskever, and Hinton}{Krizhevsky
  et~al\mbox{.}}{2012b}]%
        {krizhevsky2012imagenet}
\bibfield{author}{\bibinfo{person}{Alex Krizhevsky}, \bibinfo{person}{Ilya
  Sutskever}, {and} \bibinfo{person}{Geoffrey~E Hinton}.}
  \bibinfo{year}{2012}\natexlab{b}.
\newblock \showarticletitle{Imagenet classification with deep convolutional
  neural networks}. In \bibinfo{booktitle}{\emph{Advances in neural information
  processing systems}}. \bibinfo{pages}{1097--1105}.
\newblock


\bibitem[\protect\citeauthoryear{LeCun, Bottou, Bengio, and Haffner}{LeCun
  et~al\mbox{.}}{1998}]%
        {lecun1998gradient}
\bibfield{author}{\bibinfo{person}{Yann LeCun}, \bibinfo{person}{L{\'e}on
  Bottou}, \bibinfo{person}{Yoshua Bengio}, {and} \bibinfo{person}{Patrick
  Haffner}.} \bibinfo{year}{1998}\natexlab{}.
\newblock \showarticletitle{Gradient-based learning applied to document
  recognition}.
\newblock \bibinfo{journal}{\emph{Proc. IEEE}} \bibinfo{volume}{86},
  \bibinfo{number}{11} (\bibinfo{year}{1998}), \bibinfo{pages}{2278--2324}.
\newblock


\bibitem[\protect\citeauthoryear{Luong, Pham, and Manning}{Luong
  et~al\mbox{.}}{2015}]%
        {luong2015effective}
\bibfield{author}{\bibinfo{person}{Minh-Thang Luong}, \bibinfo{person}{Hieu
  Pham}, {and} \bibinfo{person}{Christopher~D Manning}.}
  \bibinfo{year}{2015}\natexlab{}.
\newblock \showarticletitle{Effective approaches to attention-based neural
  machine translation}.
\newblock \bibinfo{journal}{\emph{Conference on Empirical Methods in Natural
  Language Processing}} (\bibinfo{year}{2015}).
\newblock


\bibitem[\protect\citeauthoryear{Niepert, Ahmed, and Kutzkov}{Niepert
  et~al\mbox{.}}{2016}]%
        {niepert2016learning}
\bibfield{author}{\bibinfo{person}{Mathias Niepert}, \bibinfo{person}{Mohamed
  Ahmed}, {and} \bibinfo{person}{Konstantin Kutzkov}.}
  \bibinfo{year}{2016}\natexlab{}.
\newblock \showarticletitle{Learning convolutional neural networks for graphs}.
  In \bibinfo{booktitle}{\emph{International Conference on Machine Learning}}.
  \bibinfo{pages}{2014--2023}.
\newblock


\bibitem[\protect\citeauthoryear{Perozzi, Al-Rfou, and Skiena}{Perozzi
  et~al\mbox{.}}{2014}]%
        {perozzi2014deepwalk}
\bibfield{author}{\bibinfo{person}{Bryan Perozzi}, \bibinfo{person}{Rami
  Al-Rfou}, {and} \bibinfo{person}{Steven Skiena}.}
  \bibinfo{year}{2014}\natexlab{}.
\newblock \showarticletitle{Deepwalk: Online learning of social
  representations}. In \bibinfo{booktitle}{\emph{Proceedings of the 20th ACM
  SIGKDD international conference on Knowledge discovery and data mining}}.
  ACM, \bibinfo{pages}{701--710}.
\newblock


\bibitem[\protect\citeauthoryear{Ren, He, Girshick, and Sun}{Ren
  et~al\mbox{.}}{2015}]%
        {ren2015faster}
\bibfield{author}{\bibinfo{person}{Shaoqing Ren}, \bibinfo{person}{Kaiming He},
  \bibinfo{person}{Ross Girshick}, {and} \bibinfo{person}{Jian Sun}.}
  \bibinfo{year}{2015}\natexlab{}.
\newblock \showarticletitle{Faster R-CNN: Towards real-time object detection
  with region proposal networks}. In \bibinfo{booktitle}{\emph{Advances in
  neural information processing systems}}. \bibinfo{pages}{91--99}.
\newblock


\bibitem[\protect\citeauthoryear{Sen, Namata, Bilgic, Getoor, Galligher, and
  Eliassi-Rad}{Sen et~al\mbox{.}}{2008}]%
        {sen2008collective}
\bibfield{author}{\bibinfo{person}{Prithviraj Sen}, \bibinfo{person}{Galileo
  Namata}, \bibinfo{person}{Mustafa Bilgic}, \bibinfo{person}{Lise Getoor},
  \bibinfo{person}{Brian Galligher}, {and} \bibinfo{person}{Tina Eliassi-Rad}.}
  \bibinfo{year}{2008}\natexlab{}.
\newblock \showarticletitle{Collective classification in network data}.
\newblock \bibinfo{journal}{\emph{AI magazine}} \bibinfo{volume}{29},
  \bibinfo{number}{3} (\bibinfo{year}{2008}), \bibinfo{pages}{93}.
\newblock


\bibitem[\protect\citeauthoryear{Simonyan and Zisserman}{Simonyan and
  Zisserman}{2015}]%
        {simonyan2014very}
\bibfield{author}{\bibinfo{person}{Karen Simonyan} {and}
  \bibinfo{person}{Andrew Zisserman}.} \bibinfo{year}{2015}\natexlab{}.
\newblock \showarticletitle{Very deep convolutional networks for large-scale
  image recognition}. In \bibinfo{booktitle}{\emph{Proceedings of the
  International Conference on Learning Representations}}.
\newblock


\bibitem[\protect\citeauthoryear{Srivastava, Hinton, Krizhevsky, Sutskever, and
  Salakhutdinov}{Srivastava et~al\mbox{.}}{2014}]%
        {srivastava2014dropout}
\bibfield{author}{\bibinfo{person}{Nitish Srivastava},
  \bibinfo{person}{Geoffrey Hinton}, \bibinfo{person}{Alex Krizhevsky},
  \bibinfo{person}{Ilya Sutskever}, {and} \bibinfo{person}{Ruslan
  Salakhutdinov}.} \bibinfo{year}{2014}\natexlab{}.
\newblock \showarticletitle{Dropout: A simple way to prevent neural networks
  from overfitting}.
\newblock \bibinfo{journal}{\emph{Journal of Machine Learning Research}}
  \bibinfo{volume}{15}, \bibinfo{number}{1} (\bibinfo{year}{2014}),
  \bibinfo{pages}{1929--1958}.
\newblock


\bibitem[\protect\citeauthoryear{Szegedy, Liu, Jia, Sermanet, Reed, Anguelov,
  Erhan, Vanhoucke, and Rabinovich}{Szegedy et~al\mbox{.}}{2015}]%
        {szegedy2015going}
\bibfield{author}{\bibinfo{person}{Christian Szegedy}, \bibinfo{person}{Wei
  Liu}, \bibinfo{person}{Yangqing Jia}, \bibinfo{person}{Pierre Sermanet},
  \bibinfo{person}{Scott Reed}, \bibinfo{person}{Dragomir Anguelov},
  \bibinfo{person}{Dumitru Erhan}, \bibinfo{person}{Vincent Vanhoucke}, {and}
  \bibinfo{person}{Andrew Rabinovich}.} \bibinfo{year}{2015}\natexlab{}.
\newblock \showarticletitle{Going Deeper With Convolutions}. In
  \bibinfo{booktitle}{\emph{Proceedings of the IEEE Conference on Computer
  Vision and Pattern Recognition}}. \bibinfo{pages}{1--9}.
\newblock


\bibitem[\protect\citeauthoryear{Vaswani, Shazeer, Parmar, Uszkoreit, Jones,
  Gomez, Kaiser, and Polosukhin}{Vaswani et~al\mbox{.}}{2017}]%
        {vaswani2017attention}
\bibfield{author}{\bibinfo{person}{Ashish Vaswani}, \bibinfo{person}{Noam
  Shazeer}, \bibinfo{person}{Niki Parmar}, \bibinfo{person}{Jakob Uszkoreit},
  \bibinfo{person}{Llion Jones}, \bibinfo{person}{Aidan~N Gomez},
  \bibinfo{person}{{\L}ukasz Kaiser}, {and} \bibinfo{person}{Illia
  Polosukhin}.} \bibinfo{year}{2017}\natexlab{}.
\newblock \showarticletitle{Attention is all you need}. In
  \bibinfo{booktitle}{\emph{Advances in Neural Information Processing
  Systems}}. \bibinfo{pages}{6000--6010}.
\newblock


\bibitem[\protect\citeauthoryear{Veli{\v{c}}kovi{\'c}, Cucurull, Casanova,
  Romero, Li{\`o}, and Bengio}{Veli{\v{c}}kovi{\'c} et~al\mbox{.}}{2017}]%
        {velivckovic2017graph}
\bibfield{author}{\bibinfo{person}{Petar Veli{\v{c}}kovi{\'c}},
  \bibinfo{person}{Guillem Cucurull}, \bibinfo{person}{Arantxa Casanova},
  \bibinfo{person}{Adriana Romero}, \bibinfo{person}{Pietro Li{\`o}}, {and}
  \bibinfo{person}{Yoshua Bengio}.} \bibinfo{year}{2017}\natexlab{}.
\newblock \showarticletitle{Graph Attention Networks}.
\newblock \bibinfo{journal}{\emph{arXiv preprint arXiv:1710.10903}}
  (\bibinfo{year}{2017}).
\newblock


\bibitem[\protect\citeauthoryear{Wang, Wu, Coates, and Ng}{Wang
  et~al\mbox{.}}{2012}]%
        {wang2012end}
\bibfield{author}{\bibinfo{person}{Tao Wang}, \bibinfo{person}{David~J Wu},
  \bibinfo{person}{Adam Coates}, {and} \bibinfo{person}{Andrew~Y Ng}.}
  \bibinfo{year}{2012}\natexlab{}.
\newblock \showarticletitle{End-to-end text recognition with convolutional
  neural networks}. In \bibinfo{booktitle}{\emph{Pattern Recognition (ICPR),
  2012 21st International Conference on}}. IEEE, \bibinfo{pages}{3304--3308}.
\newblock


\bibitem[\protect\citeauthoryear{Yang, Cohen, and Salakhutdinov}{Yang
  et~al\mbox{.}}{2016}]%
        {yang2016revisiting}
\bibfield{author}{\bibinfo{person}{Zhilin Yang}, \bibinfo{person}{William~W
  Cohen}, {and} \bibinfo{person}{Ruslan Salakhutdinov}.}
  \bibinfo{year}{2016}\natexlab{}.
\newblock \showarticletitle{Revisiting semi-supervised learning with graph
  embeddings}.
\newblock \bibinfo{journal}{\emph{International Conference on Machine
  Learning}} (\bibinfo{year}{2016}).
\newblock


\bibitem[\protect\citeauthoryear{Zitnik and Leskovec}{Zitnik and
  Leskovec}{2017}]%
        {zitnik2017predicting}
\bibfield{author}{\bibinfo{person}{Marinka Zitnik} {and} \bibinfo{person}{Jure
  Leskovec}.} \bibinfo{year}{2017}\natexlab{}.
\newblock \showarticletitle{Predicting multicellular function through
  multi-layer tissue networks}.
\newblock \bibinfo{journal}{\emph{Bioinformatics}} \bibinfo{volume}{33},
  \bibinfo{number}{14} (\bibinfo{year}{2017}), \bibinfo{pages}{i190--i198}.
\newblock


\end{thebibliography}

\end{document}